\documentclass[11pt,a4paper]{article}
\usepackage[hyperref]{acl2017}
\usepackage{times}
\usepackage{latexsym}

\usepackage{url}
\usepackage{amssymb,amsfonts}
\usepackage{multirow}
\usepackage{arydshln}
\usepackage{amsmath}
\usepackage{rotating}
\usepackage{color}
\usepackage{subfigure}

\aclfinalcopy % Uncomment this line for the final submission
\def\figref#1{Figure~\ref{fig:#1}}

\def\tabref#1{Table~\ref{tab:#1}}

\def\seclabel#1{\label{sec:#1}\label{p:#1}}
\def\eqref#1{Eq.~\ref{eqn:#1}}

\long\def\eat#1{\ignorespaces}

\title{Comparative Study of CNN and RNN for Natural Language Processing}

\author{Wenpeng Yin$^{\dagger}$, Katharina Kann$^{\dagger}$, Mo Yu$^{\ddagger}$ \and Hinrich Sch\"utze$^{\dagger}$\\
  {\tt $^{\dagger}$CIS, LMU Munich, Germany}\\
  {\tt $^{\ddagger}$IBM Research, USA}\\
{\tt \small \{wenpeng,kann\}@cis.lmu.de, yum@us.ibm.com}
}

\date{}

\newcounter{notecounter}

\newcommand{\enoteson}{\long\gdef\enote##1##2{{
\stepcounter{notecounter}
\large\bf
\hspace{1cm}\arabic{notecounter} $<<<$ ##1: ##2
$>>>$\hspace{1cm}}}}
\enoteson
\begin{document}
\maketitle

\begin{abstract}
Deep neural networks (DNNs) have revolutionized the field of
natural language processing (NLP). Convolutional Neural
Network (CNN) and Recurrent Neural Network (RNN), the two
main types of DNN architectures,
are widely explored to handle
various NLP tasks. CNN is supposed to be good at extracting
position-invariant features and RNN at
modeling units in sequence. The state-of-the-art on many NLP
tasks often switches due to the battle of CNNs and RNNs. This
work is the first systematic comparison of CNN and RNN on a wide range of  representative NLP tasks,
aiming to give basic guidance for DNN selection.
\end{abstract}

\section{Introduction}
\seclabel{intro} 

Natural language processing (NLP) has
benefited greatly from the resurgence of deep neural
networks (DNNs), due to their high performance with less need of
engineered features. There are two main
DNN architectures:
convolutional neural network (CNN)
\cite{lecun1998gradient} and recurrent neural
network (RNN) \cite{journalsElman90}. Gating mechanisms have been
developed to alleviate some limitations of the basic RNN,
resulting in two
prevailing RNN types: long short-term
memory (LSTM) \cite{hochreiter1997long} and gated
recurrent unit (GRU) \cite{cho2014properties}.

Generally speaking, CNNs are hierarchical and RNNs
sequential architectures. How should we choose between
them for
processing language?  Based on the  characterization
``hierarchical (CNN) vs.\  sequential (RNN)'', it is
tempting
to choose a CNN for classification tasks like
sentiment classification since sentiment is usually
determined by some key phrases; and to choose RNNs for a
sequence modeling task like language modeling as it
requires flexible modeling of context dependencies.
But current NLP literature does not support such a clear conclusion.
For example,
RNNs
perform well on document-level sentiment classification
\cite{tang2015document};
and \newcite{dauphin2016language} recently showed
that gated CNNs outperform LSTMs on language
modeling tasks, even though
LSTMs had long been seen as better suited.
In summary, there is no consensus on DNN selection for any particular NLP problem.

This work compares CNNs, GRUs and LSTMs
systematically on a broad array of NLP tasks:
sentiment/relation classification, textual entailment,
answer selection, question-relation matching in Freebase, Freebase path query answering and
part-of-speech  tagging.

Our experiments support two key findings.
(i) CNNs and RNNs provide complementary
information for text classification tasks. Which architecture performs better
depends on how important it is to \emph{semantically understand the
whole sequence}. (ii) Learning rate changes
performance relatively smoothly, while changes to hidden size and
batch size result in large fluctuations.

\begin{figure*}[t] 
\centering 
\subfigure[CNN] { \label{fig:cnn} 
\includegraphics[width=5cm]{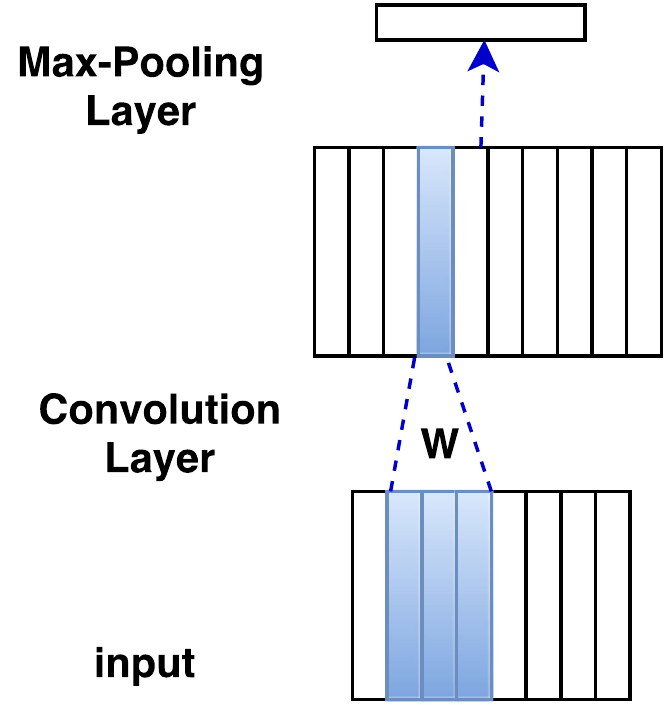} 
} 
\subfigure[GRU] { \label{fig:gru} 
\includegraphics[width=5cm]{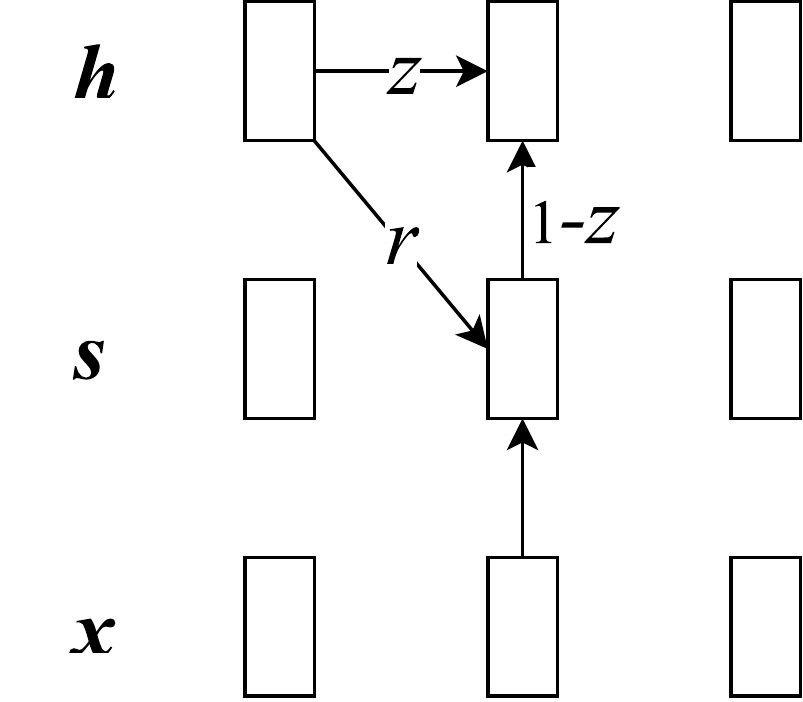} 
} 
\subfigure[LSTM] { \label{fig:lstm} 
\includegraphics[width=5cm]{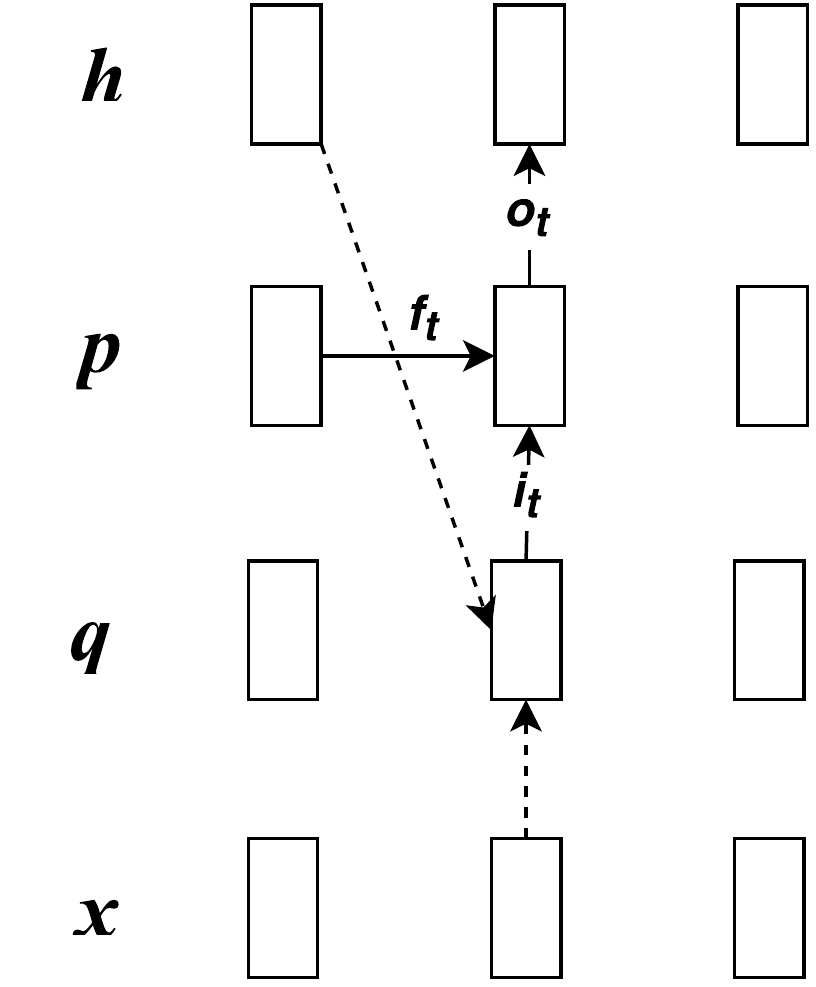} 
} 
\caption{Three typical DNN architectures} 
\label{fig:nns} 
\end{figure*}
\section{Related Work}

To our knowledge, there has been no systematic comparison of CNN and RNN on a
large array of NLP tasks.

\newcite{VuAGS16} investigate  CNN and basic RNN (i.e., no
gating mechanisms) for relation classification. They report
higher performance of CNN than RNN and give evidence that  CNN and RNN provide
complementary information: while the RNN computes
a weighted combination of all words in the
sentence, the CNN extracts the most informative ngrams
for the relation and only considers their resulting
activations. Both \newcite{DBLPWenZLW16} and
\newcite{heike2017} support CNN over GRU/LSTM  for classification  of long sentences. In addition, \newcite{YinSXZ16} achieve better performance of attention-based CNN than attention-based LSTM for answer
selection. \newcite{dauphin2016language} further argue
that a fine-tuned gated CNN can also model long-context
dependency, getting new state-of-the-art in language
modeling above all RNN competitors

In contrast, \newcite{Arkhipenko2016} compare word2vec \cite{DBLPMikolovSCCD13}, CNN, GRU and LSTM in sentiment analysis of Russian tweets, and find GRU outperforms LSTM and CNN.

In empirical evaluations,
\newcite{chung2014empirical} and
\newcite{jozefowicz2015empirical} found there is
no  clear winner between GRU and LSTM. In many tasks, they yield
comparable performance and tuning hyperparameters like layer
size is often more important than picking the ideal
architecture.

\section{Models}
This section  gives a brief introduction of CNN, GRU and LSTM.

\subsection{Convolutional Neural Network (CNN)}

\paragraph{Input Layer} Sequence $x$ contains $n$ entries. Each
entry is represented by
a $d$-dimensional dense vector; thus
the input $x$ is represented as a feature map of
dimensionality  $d \times n$. 
Figure \ref{fig:cnn} shows the input layer as
the lower rectangle with multiple columns.

\paragraph{Convolution Layer}  is used for  representation
learning from sliding $w$-grams. For an input sequence with $n$ entries: $x_1,x_2,\ldots,x_n$,
let vector $\mathbf{c}_i\in\mathbb{R}^{wd}$ be the
concatenated embeddings of $w$ entries
$x_{i-w+1},\ldots,x_{i}$ where $w$ is the filter width and $0< i
<s+w$.  Embeddings for  $x_i$, $i<1$ or $i>n$, are
zero padded.  We then generate the representation
$\mathbf{p}_i\in\mathbb{R}^d$ for the $w$-gram
$x_{i-w+1},\ldots,x_{i}$ using the convolution weights
$\mathbf{W}\in\mathbb{R}^{d\times wd}$:
\begin{equation}\label{eq:cnn}
\mathbf{p}_i=\mathrm{tanh}(\mathbf{W}\cdot\mathbf{c}_i+\mathbf{b})
\end{equation}
where bias $\mathbf{b}\in\mathbb{R}^d$. 

\paragraph{Maxpooling}
All $w$-gram representations $\mathbf{p}_i$ $(i=1\cdots s+w-1)$ are used to generate the  representation of input sequence $x$ by maxpooling: $\mathbf{x}_j=\mathrm{max}(\mathbf{p}_{1,j}, \mathbf{p}_{2,j}, \cdots)$ ($j=1,\cdots, d$).

\subsection{Gated Recurrent Unit (GRU)}
GRU, as shown in Figure \ref{fig:gru}, models text $x$ as follows:
\begin{eqnarray}\label{equ:gru}
%\setlength{\abovedisplayskip}{-8pt}
%\setlength{\belowdisplayskip}{3pt}
%\begin{split}
\mathbf{z}&=&\sigma(\mathbf{x}_t\mathbf{U}^z+\mathbf{h}_{t-1}\mathbf{W}^z)\\
\mathbf{r}&=&\sigma(\mathbf{x}_t\mathbf{U}^r+\mathbf{h}_{t-1}\mathbf{W}^r)\\
\mathbf{s}_t&=&\mathrm{tanh}(\mathbf{x}_t\mathbf{U}^s+(\mathbf{h}_{t-1}\circ \mathbf{r})\mathbf{W}^s)\\
\mathbf{h}_t&=&(1-\mathbf{z})\circ \mathbf{s}_t+\mathbf{z}\circ \mathbf{h}_{t-1}
%\end{split}
\end{eqnarray}
$\mathbf{x}_t\in\mathbb{R}^d$  represents the token in $x$ at  position $t$,
$\mathbf{h}_t\in\mathbb{R}^h$ is the hidden state at  $t$, supposed  to encode the history $x_1$, $\cdots$, $x_t$. $\mathbf{z}$ and $\mathbf{r}$ are two gates. All $\mathbf{U}\in\mathbb{R}^{d\times h}$,$\mathbf{W}\in\mathbb{R}^{h\times h}$ are parameters.

\subsection{Long Short-Time Memory (LSTM)}

LSTM is denoted in Figure \ref{fig:lstm}. It models the word sequence $x$ as follows:
\begin{eqnarray}\label{equ:lstm}
%\setlength{\abovedisplayskip}{-8pt}
%\setlength{\belowdisplayskip}{3pt}
%\begin{split}
\mathbf{i}_t&=&\sigma(\mathbf{x}_t\mathbf{U}^i+\mathbf{h}_{t-1}\mathbf{W}^i+\mathbf{b}_i)\\
\mathbf{f}_t&=&\sigma(\mathbf{x}_t\mathbf{U}^f+\mathbf{h}_{t-1}\mathbf{W}^f+\mathbf{b}_f)\\
\mathbf{o}_t&=&\sigma(\mathbf{x}_t\mathbf{U}^o+\mathbf{h}_{t-1}\mathbf{W}^o+\mathbf{b}_o)\\
\mathbf{q}_t&=&\mathrm{tanh}(\mathbf{x}_t\mathbf{U}^q+\mathbf{h}_{t-1}\mathbf{W}^q+\mathbf{b}_q)\\
\mathbf{p}_t &=& \mathbf{f}_t*\mathbf{p}_{t-1} + \mathbf{i}_t * \mathbf{q}_t\\
\mathbf{h}_t &=& \mathbf{o}_t * \mathrm{tanh}(\mathbf{p}_t)
\end{eqnarray}

LSTM has three gates: input gate $i_t$, forget gate $f_t$ and output gate $o_t$. All gates are generated by a \emph{sigmoid} function over the ensemble of input $x_t$ and the preceding hidden state $h_{t-1}$. In order to generate the hidden state at current step $t$, it first generates a temporary result $q_t$ by a \emph{tanh} non-linearity over the ensemble of input $x_t$ and the preceding hidden state $h_{t-1}$, then combines this temporary result $q_t$ with history $p_{t-1}$ by input gate $i_t$ and forget gate $f_t$ respectively to get an updated history $p_t$, finally uses output gate $o_t$ over this updated history $p_t$ to get the final hidden state $h_t$.

\section{Experiments}
\begin{table*}[t]
\setlength{\tabcolsep}{2pt}
%\small
  \centering
  \begin{tabular}{c c r|c|c c c c c c c}\hline
    %\textbf{Systems}& \textbf{ROUGE-1} & \textbf{ROUGE-2} & \textbf{ROUGE-SU4}\\ \hline \hline
   & &&performance & lr & hidden & batch & sentLen & filter\_size &margin\\ \hline \hline
  \multirow{6}{*}{TextC} &  \multirow{3}{*}{SentiC (acc)}& CNN&82.38 &0.2  & 20 & 5 & 60 &3&--\\
&&GRU    &\textbf{86.32} &0.1  & 30 & 50 & 60 &--&--\\
&&LSTM    &84.51 &0.2 & 20 & 40 & 60&--&--\\\cline{2-10}

 &   \multirow{3}{*}{RC (F1)}& CNN& 68.02 &0.12 & 70 & 10 & 20 &3&--\\
&&GRU    &\textbf{68.56} & 0.12 & 80 & 100 & 20 &--&--\\
&&LSTM    &66.45 &0.1 &  80& 20 &20 &--&--\\\hline\hline

  \multirow{9}{*}{SemMatch} &  \multirow{3}{*}{TE (acc)}& CNN&77.13 &0.1 & 70 & 50 & 50 &3&--\\
&&GRU    &\textbf{78.78} & 0.1 & 50 & 80& 65 & --&--\\
&&LSTM    &77.85 &0.1 & 80 & 50 &50 &--&--\\\cline{2-10}

  &  \multirow{3}{*}{AS (MAP \& MRR)}& CNN&  (\textbf{63.69,65.01}) & 0.01 & 30 & 60 &40&3&0.3\\
&&GRU    &(62.58,63.59) &0.1 & 80 &150 & 40 & --&0.3\\
&&LSTM    &(62.00,63.26) & 0.1& 60  & 150  & 45 &--& 0.1\\\cline{2-10}

 &   \multirow{3}{*}{QRM (acc)}& CNN& \textbf{71.50}& 0.125 & 400 & 50 &17 & 5 & 0.01\\
&&GRU    &69.80 & 1.0 & 400 & 50 &17 & - &0.01 \\
&&LSTM    & 71.44& 1.0 & 200 & 50 &17 & - &0.01\\\hline\hline

 \multirow{3}{*}{SeqOrder} &   \multirow{3}{*}{PQA (hit@10)}& CNN& 54.42& 0.01  & 250 & 50 &5&3 &0.4\\
&&GRU    & \textbf{55.67}& 0.1 & 250 &50 &5  & -- & 0.3\\
&&LSTM    &55.39 & 0.1 & 300 & 50 & 5 &-- & 0.3\\\hline \hline

 \multirow{5}{*}{ContextDep} &   \multirow{5}{*}{POS tagging (acc)}& CNN& 94.18& 0.1& 100 & 10 & 60 & 5 &--\\
&&GRU    & 93.15 &0.1  & 50 & 50& 60 & --&--\\
&&LSTM    &93.18 & 0.1& 200 & 70 & 60&-- & --\\
&&Bi-GRU    & 94.26 & 0.1 & 50& 50& 60 & -- & --\\
&&Bi-LSTM    & \textbf{94.35} & 0.1 & 150 & 5 & 60 & -- & --\\\hline
  \end{tabular}
\caption{Best results or CNN, GRU and LSTM in NLP tasks}\label{tab:overallresult}
\end{table*}

\def\sentsep{0.05cm}

% \begin{table*}[t]
% % \setlength{\belowcaptionskip}{-10pt}
% % \setlength{\abovecaptionskip}{5pt}
% %\small
%   \centering
%   \begin{tabular}{c c r|c}\\\hline
  
% &&&(1) It 's a movie -- and an album -- you wo n't want to miss\\
% &F&T & F & (2) These are names to remember , in order to avoid them in the future\\\hline
% \multirow{5}{*}{\rotatebox{90}{\scriptsize GRU wrong}}&F & F & T &
% (3) In the second half of the film , Frei 's control loosens in direct proportion to the
% amount of screen time he gives Nachtwey for self-analysis\\
% &T&T & F & (4) The result is mesmerizing -- filled with menace and squalor
%   \end{tabular}
% \caption{CNN vs GRU in Sentiment Classification}\label{tab:disagreement}
% \end{table*}

\begin{table*}[t]
\setlength{\tabcolsep}{2pt}
  \centering
  \begin{tabular}{c c c |l}\hline
Gold & CNN & GRU & examples\\\hline\hline
T & F & T & It 's a movie -- and an album -- you wo n't want to miss\\
F&T & F & These are names to remember , in order to avoid them in the future\\\hline
\multirow{2}{*}{F} & \multirow{2}{*}{F} & \multirow{2}{*}{T} & In the second half of the film , Frei 's control loosens in direct proportion to the \\
 & & & amount of screen time he gives nachtwey for self-analysis\\\hdashline
T&T & F & The result is mesmerizing -- filled with menace and squalor\\\hline
  \end{tabular}
\caption{CNN vs GRU in Sentiment Classification}\label{tab:disagreement}
\end{table*}

\begin{figure*}[t]
\centering
\begin{tabular}{ll}
\includegraphics[width=6.9cm]{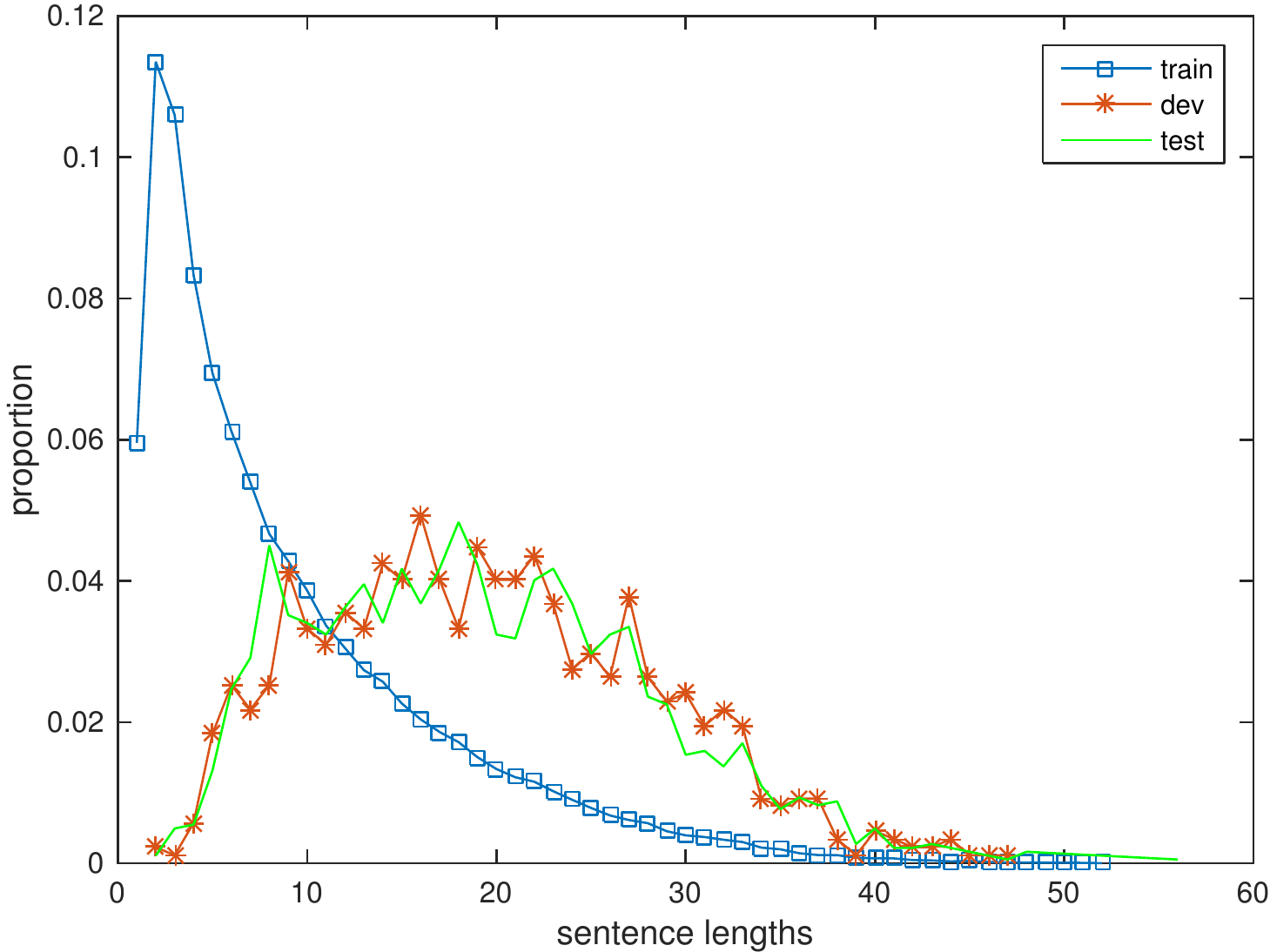} &
\includegraphics[width=6.9cm]{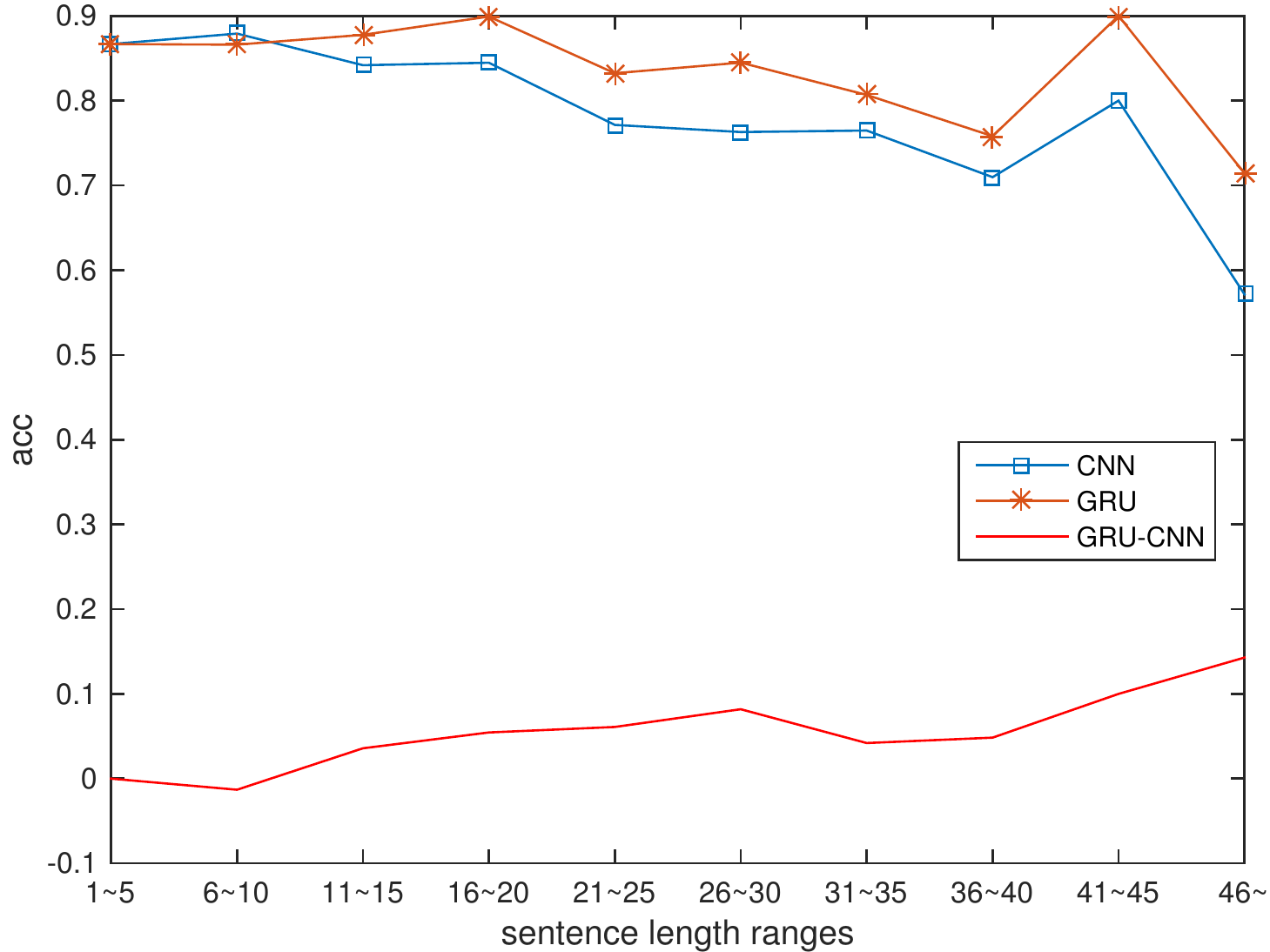}
\end{tabular}
\caption{Distributions of sentence lengths (left) and accuracies of different length ranges (right).
\label{fig:lengthsprop}
\label{fig:accvslengths}}
\end{figure*}
\begin{figure*}[t]
\centering
\begin{tabular}{lll}
\includegraphics[width=5cm]{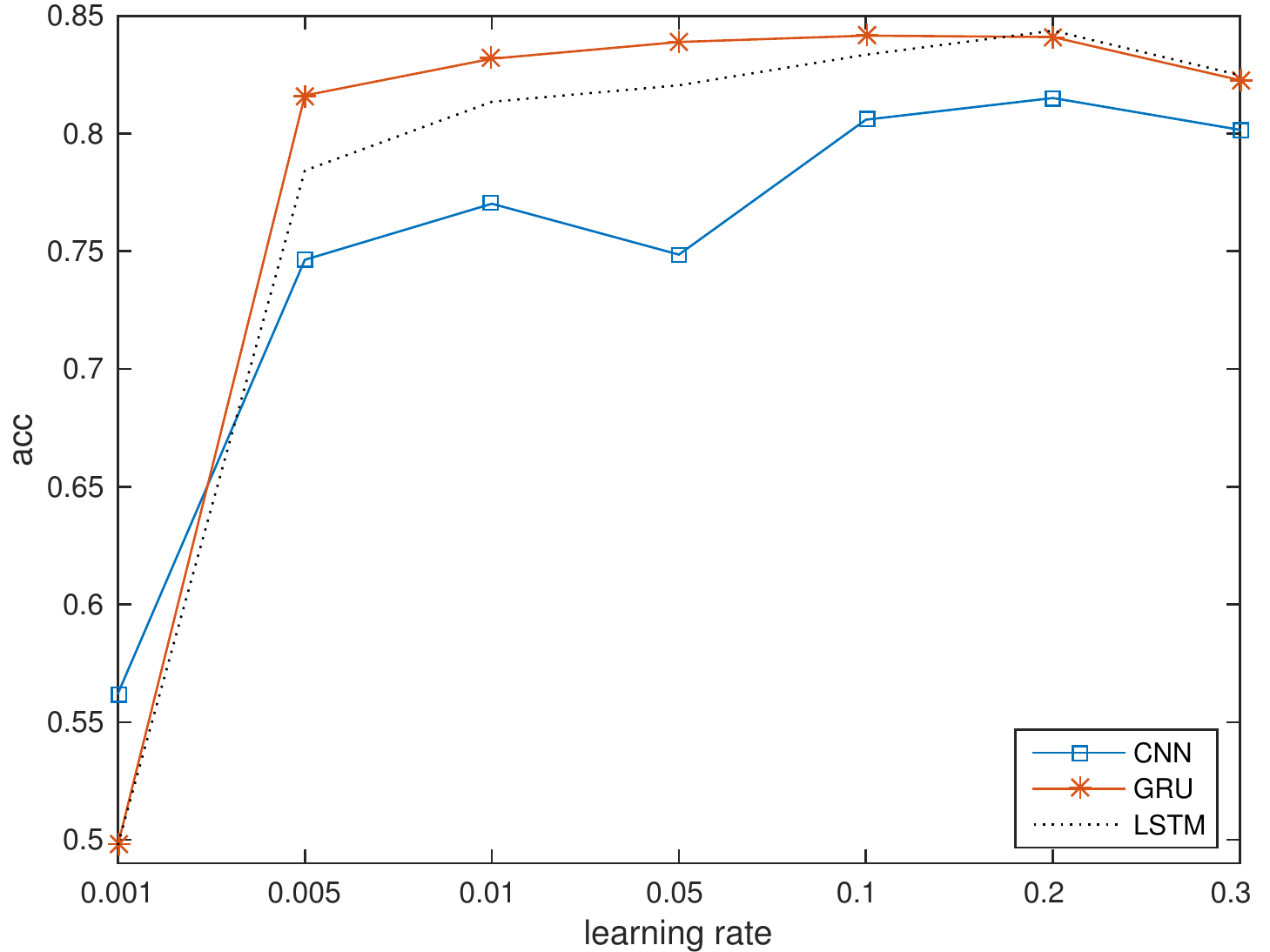} &
\includegraphics[width=5cm]{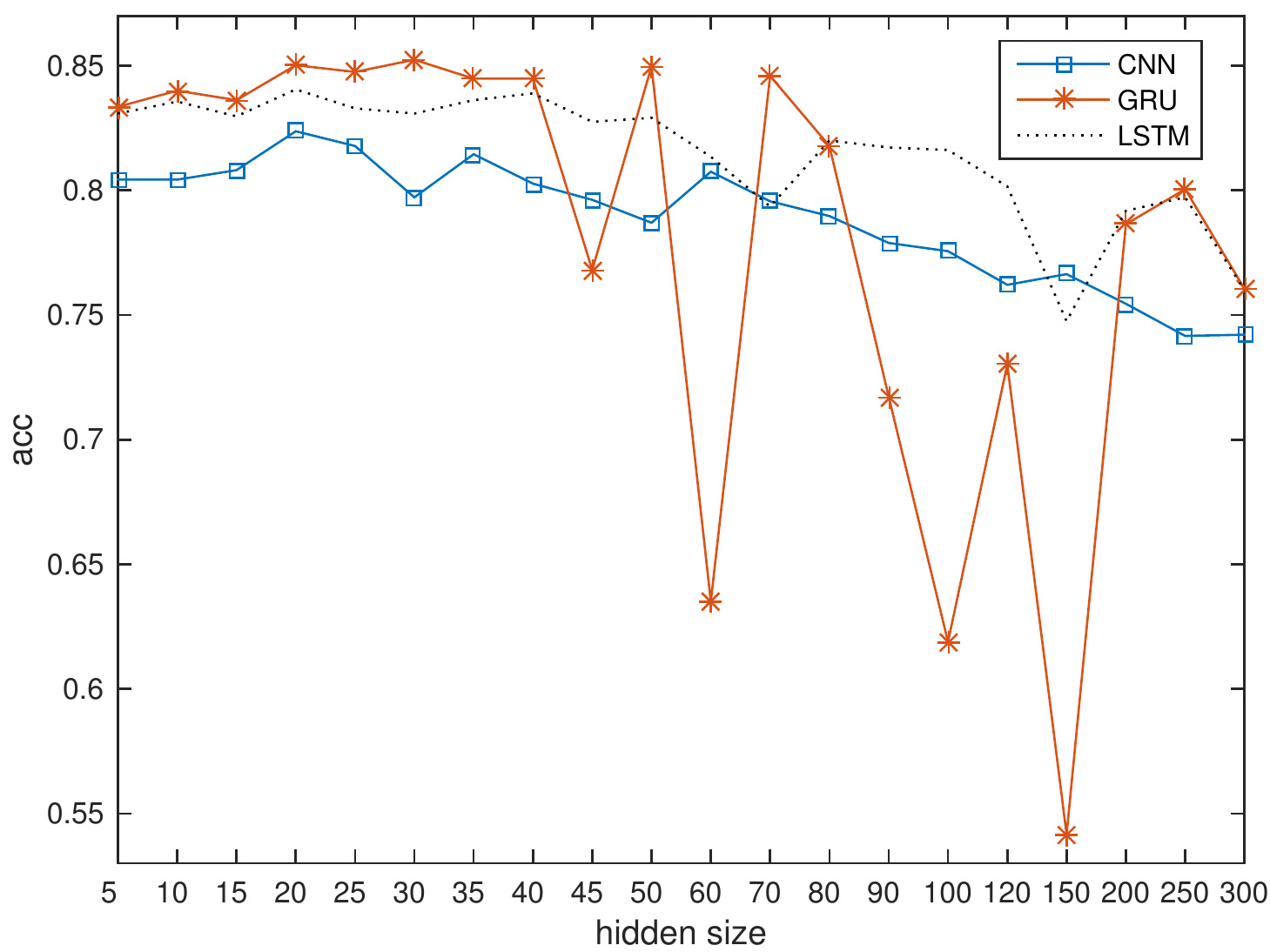} &
\includegraphics[width=5cm]{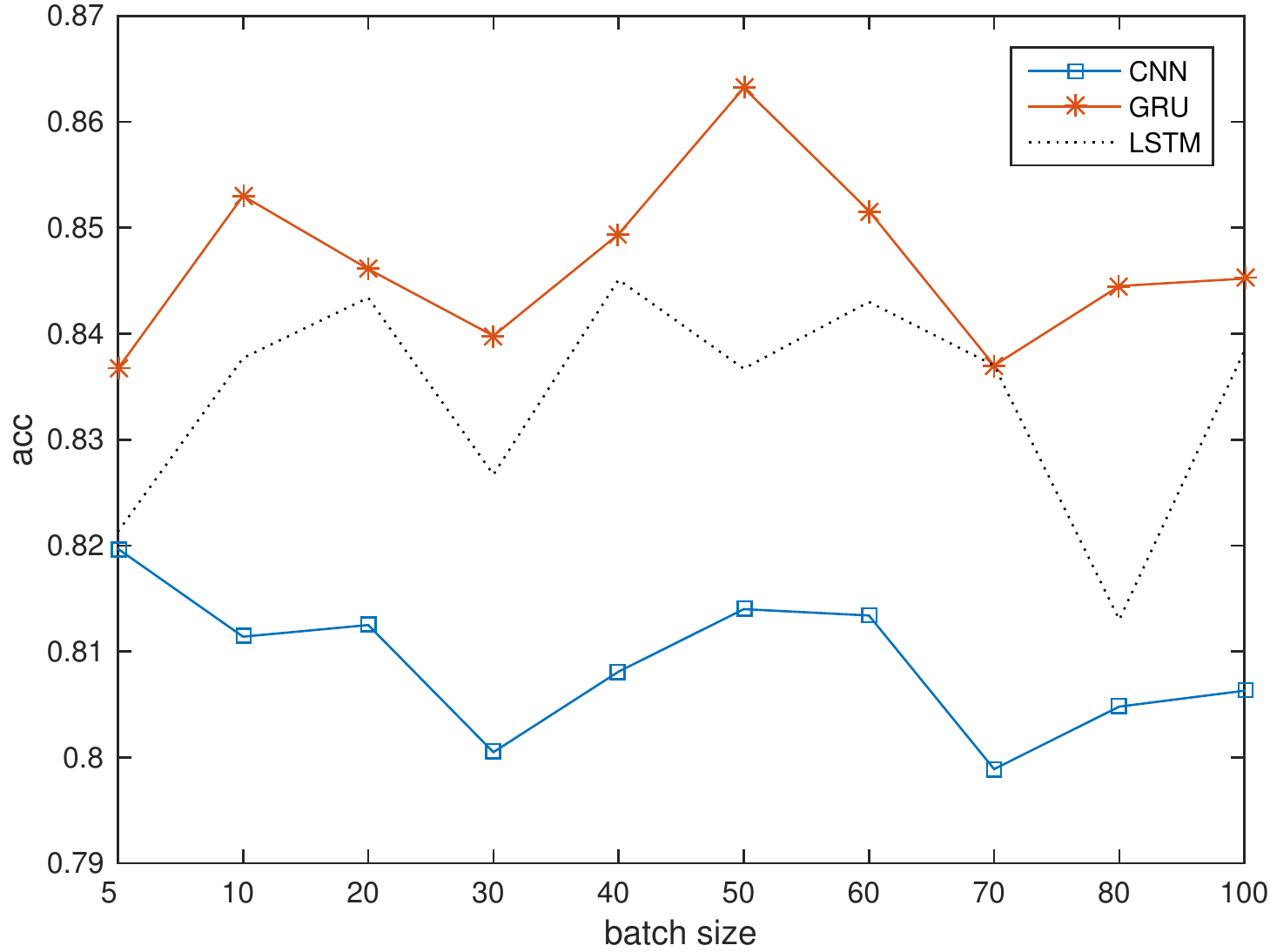} \\
\includegraphics[width=5cm]{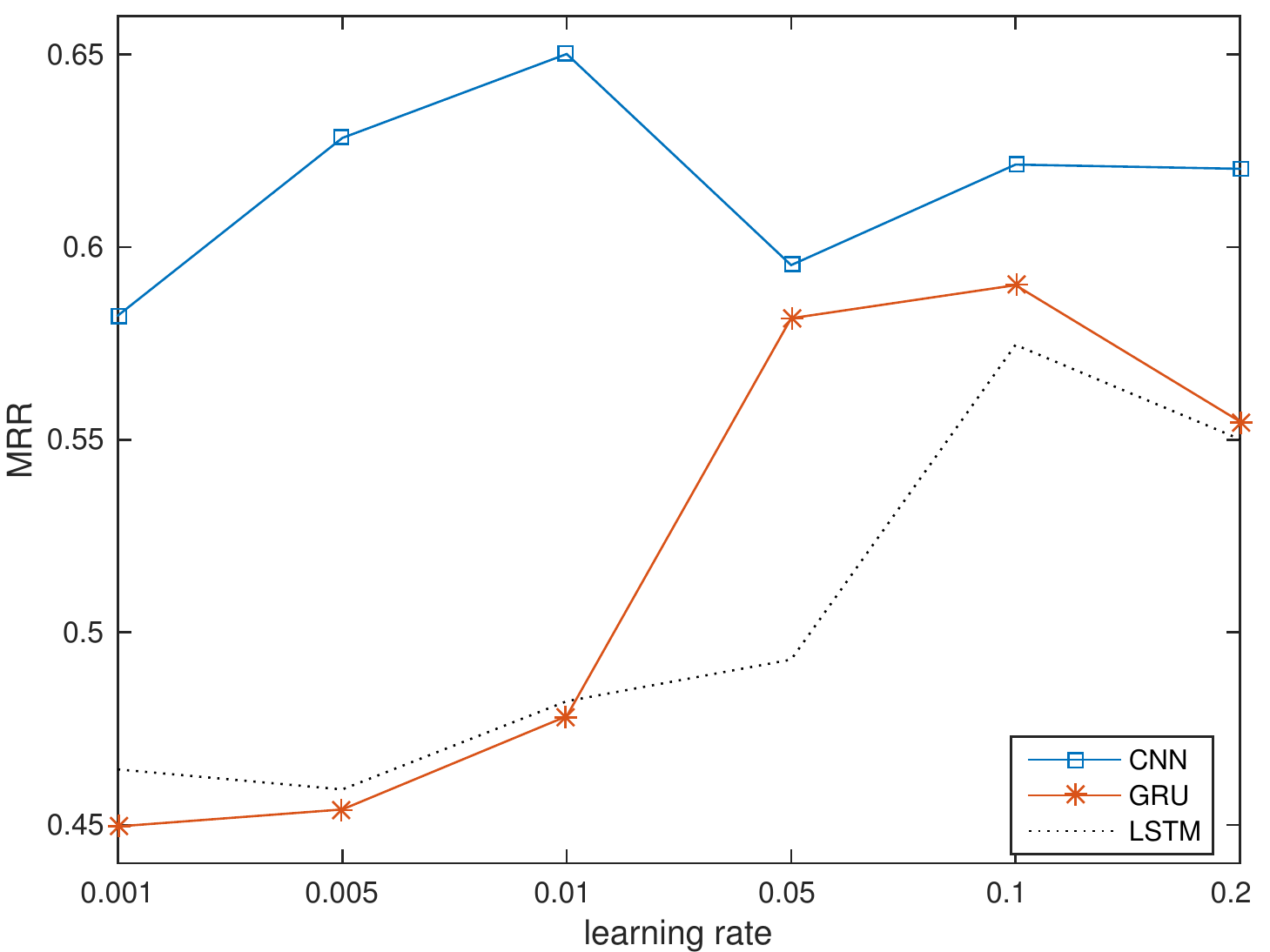} &
\includegraphics[width=5cm]{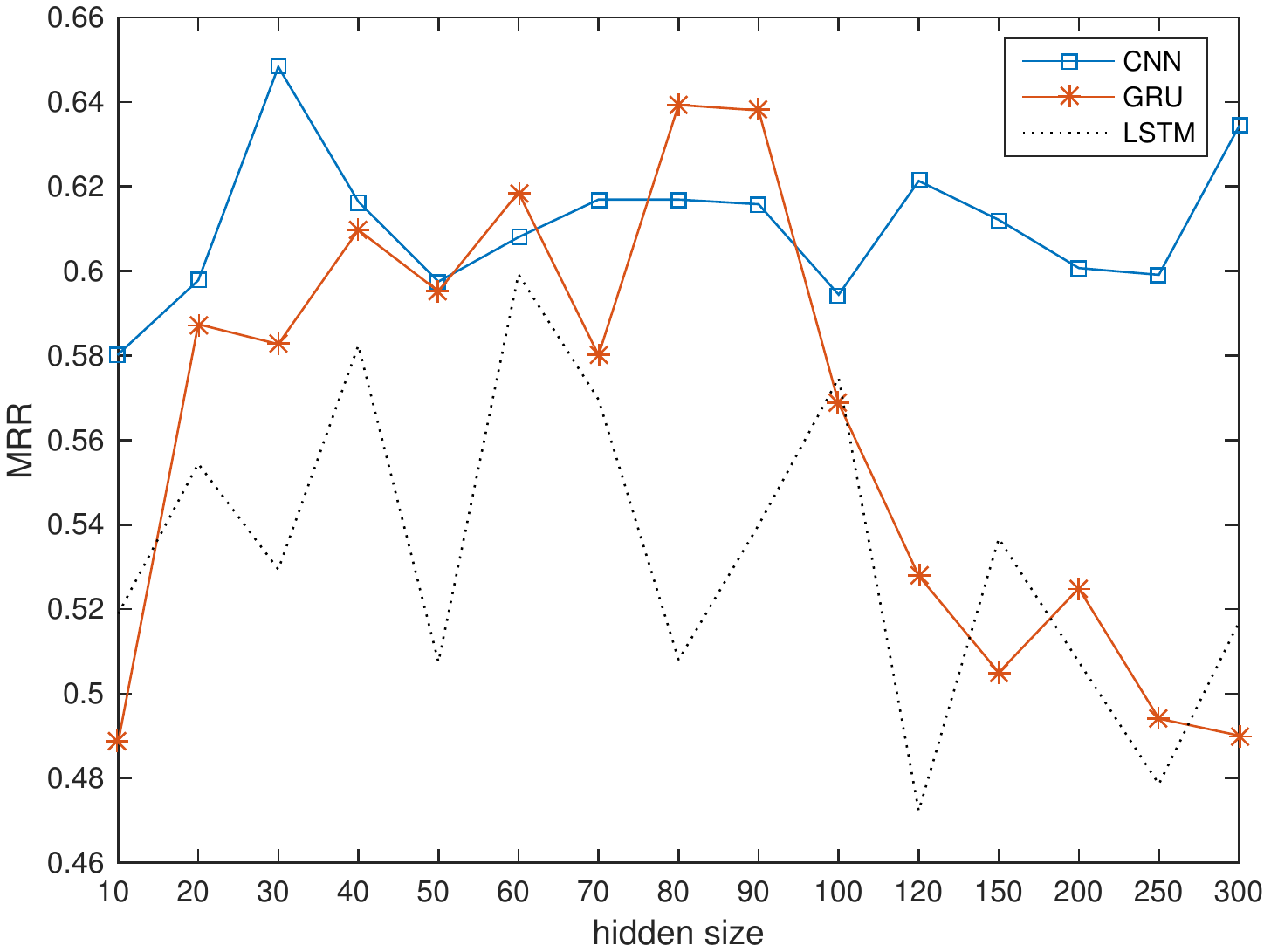} &
\includegraphics[width=5cm]{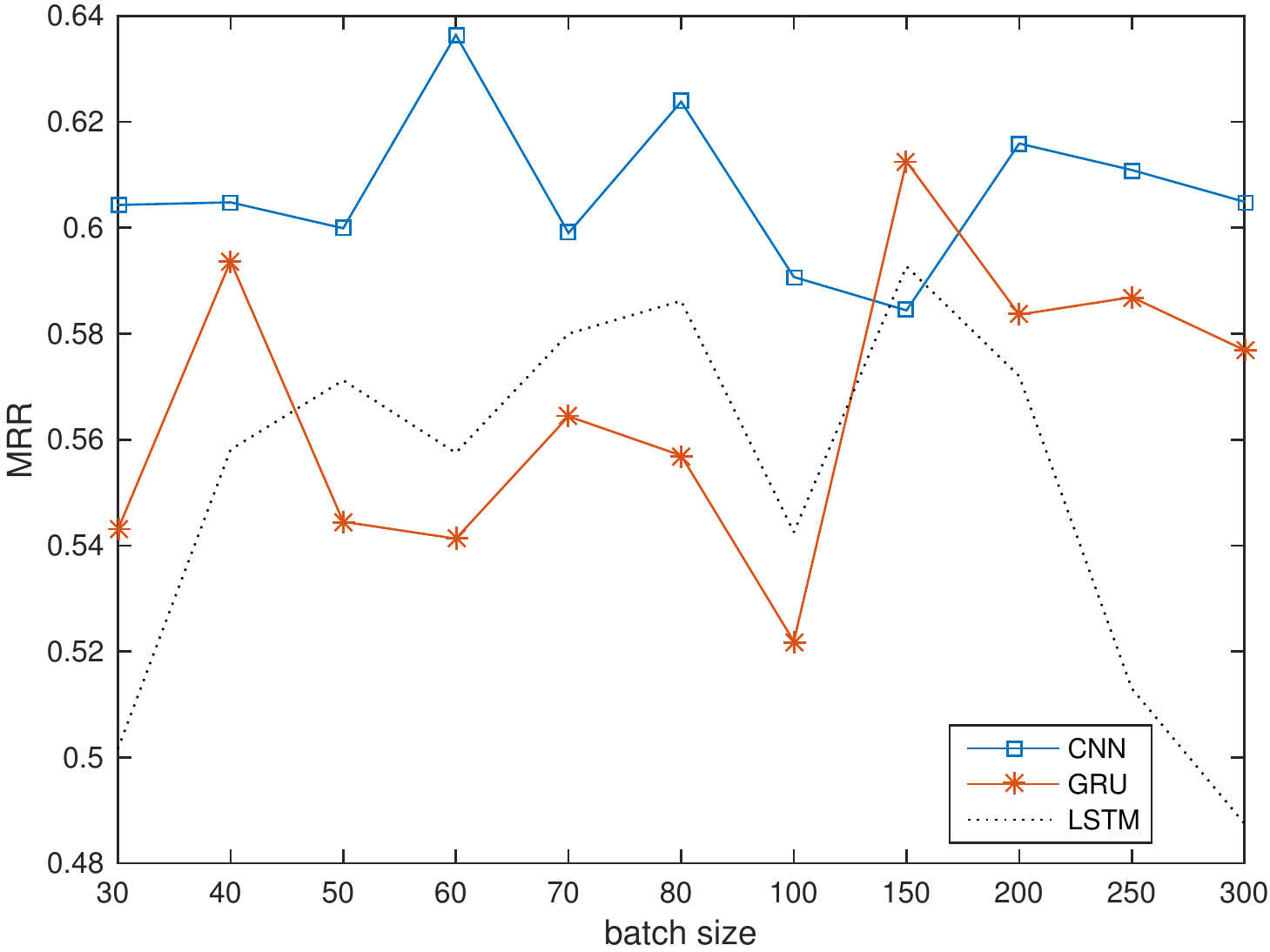}
\end{tabular}
\caption{Accuracy for sentiment classification (top) and MRR
  for WikiQA (bottom) as a function of three hyperparameters: learning rate
  (left), hidden size (center) and batch size (right).
\label{fig:wikiqa}
\label{fig:senti}}
\end{figure*}

\subsection{Tasks}
\textbf{Sentiment Classification (SentiC)} on
Stanford Sentiment Treebank (SST) \cite{socher2013recursive}. This dataset
predicts the sentiment (positive or negative) of movie reviews. We use
the given split of 6920 \emph{train}, 872 \emph{dev}
and 1821 \emph{test} sentences.
 As in
\cite{kalchbrenner2014convolutional,le2014distributed}, we treat
labeled phrases that occur
as subparts of  training sentences
as independent training instances.
Measure: accuracy.

\textbf{Relation Classification (RC)} on
SemEval 2010 task 8 \cite{hendrickx2009semeval}. It consists of sentences which have been manually labeled
with 19 relations (18 directed relations and
\emph{Other}), 8000 sentences in \emph{train} and 2717
in \emph{test}. As there is no dev set, we use
1500 training examples as \emph{dev}, similar to  \newcite{VuAGS16}. Measure: F1.

\textbf{Textual Entailment (TE)} on
Stanford Natural Language Inference (SNLI)
\cite{bowman2015large}. SNLI contains premise-hypothesis
pairs, labeled with a relation (entailment,
contradiction, neutral). After removing  unlabeled pairs, we end up having 549,367 pairs for \emph{train}, 9,842 for \emph{dev} and 9,824 for \emph{test}. Measure: accuracy.

\textbf{Answer Selection (AS)}
on WikiQA \cite{yang2015wikiqa}, an open domain
question-answer dataset. We use the subtask that assumes that
there is at least one correct answer for a question.
The corresponding dataset consists of 20,360
question-candidate pairs in \emph{train}, 1,130  in \emph{dev} and
2,352 in \emph{test} where we
adopt the standard setup
of only
considering questions with correct answers in test. The task is to choose the correct answer(s) from some candidates for a question.  Measures: MAP and MRR.

\textbf{Question Relation Match (QRM).}  We utilize WebQSP
\cite{yih-EtAl:2016:P16-2} dataset to create a large-scale
relation detection task, benefitting from the availability of
labeled semantic parses of questions.  For each question, we
(i) select the topic entity from the parse; (ii) select all
the relations/relation chains (length $\leq$ 2) connecting
to the topic entity; and (iii) set the
relations/relation-chains in the labeled parse as positive
and all the others as negative.  Following
\newcite{yih-EtAl:2016:P16-2} and \newcite{XuFRHZ16}, we formulate this task
as \emph{a sequence matching problem}.  Ranking-loss is used
for training. Measure: accuracy.

\textbf{Path Query Answering (PQA)}
on the path query dataset released by
\newcite{GuML15}. It contains KB paths like $e_h, r_0, r_1,
\cdots, r_t, e_t$, where head entity $e_h$ and relation sequence $r_0, r_1, \cdots, r_t$ are encoded to predict the tail entity
 $e_t$. There are
6,266,058/27,163/109,557 paths in \emph{train/dev/test},
respectively. Measure: hit@10.

\textbf{Part-of-Speech Tagging} on WSJ. We use the setup
of \cite{blitzer2006domain,petrov2012overview}: sections 2-21 are
\emph{train}, section 22 is \emph{dev} and section 23 is
\emph{test}. Measure: accuracy.

We organize above tasks in four categories. (i)
\textbf{TextC.}
Text
  classification, including SentiC and RC. (ii)
\textbf{SemMatch} including TE, AS and QRM. (iii)
\textbf{SeqOrder.} Sequence order, i.e., \emph{PQA}.
(iv) \textbf{ContextDep.} Context dependency, i.e., POS
  tagging.
By investigating these four categories,
we aim to discover some
basic principles involved in utilizing CNNs / RNNs.

\subsection{Experimental Setup}
To fairly study the encoding capability of different \emph{basic DNNs},
our
experiments have the following design. (i) Always train
from scratch, no  extra knowledge, e.g., no
pretrained word embeddings. (ii)
Always train using a basic setup without complex
tricks such
as batch
normalization. (iii) Search for optimal hyperparameters for
each task and each model
separately, so that all results are based on optimal hyperparameters.
(iv) Investigate the \emph{basic
architecture and utilization} of each model: CNN  consists of a
convolution layer and a max-pooling layer; GRU and LSTM
model the input from left to right and  always use the
last hidden state as the final representation of the input. An exception is for POS tagging, we also report
bi-directional RNNs as this can make sure each word's representation can encode the word's context of both sides, like
the CNN does.

Hyperparameters are tuned on dev: hidden size, minibatch size,
learning rate, maximal sentence length, filter size (for CNN
only) and margin in ranking loss in AS, QRM and PQA tasks.

%-----------------
%experiments results
%--------------

\subsection{Results \& Analysis}

\tabref{overallresult} shows experimental results
for all tasks and models and corresponding
hyperparameters. For TextC,
GRU performs best on  SentiC and
comparably with CNN in RC. For SemMatch,
CNN performs best on AS
and QRM while GRU (and also LSTM) outperforms CNN on TE.
For SeqOrder (PQA), both GRU and LSTM
outperform CNN. For ContextDep (POS tagging), CNN outperforms
one-directional RNNs, but lags behind  bi-directional
RNNs.

The results for
SeqOrder and ContextDep
are as expected: RNNs are well suited
to encode order information (for PQA) and long-range
context dependency (for POS tagging). But for the
other two categories, TextC and
SemMatch, some
unexpected observations appear.
CNNs are considered  good at extracting local and position-invariant
features and therefore should perform well on TextC;
but in our experiments they are outperformed by RNNs, especially in SentiC. How
can this be explained?
RNNs can encode the structure-dependent semantics of the
whole input, but how likely is this helpful for TextC
tasks that mostly depend on a few local regions? To
investigate the unexpected observations, we do some error analysis on
SentiC.

\paragraph{Qualitative analysis}
\tabref{disagreement} shows examples (1) -- (4)
in which CNN predicts correctly while GRU predicts falsely
or vice versa. We find that GRU is better when
sentiment is determined by the entire sentence or a
long-range semantic
dependency --
rather than some local key-phrases -- is involved.
Example (1)
contains the phrases ``won't'' and
``miss'' that usually appear with negative sentiment,
but the whole sentence describes a positive sentiment;
thus, an architecture like  GRU is needed that handles long
sequences correctly.
On the other hand, modeling the whole sentence sometimes is a burden --
neglecting the key parts. The
GRU encodes the entire word sequence of the long example
(3), making it hard for the
negative keyphrase ``loosens'' to
play a main role in the final representation. The first part
of example (4)
seems
positive while the second part seems negative. As GRU
chooses the last hidden state to represent the sentence,
this might result in the wrong prediction.

Studying acc vs sentence length can also support this. \figref{accvslengths} (left) shows sentence lengths in SST are mostly short in \emph{train} while close to normal distribution around 20 in \emph{dev} and \emph{test}. \figref{accvslengths} (right) depicts the accuracies w.r.t length ranges. We found that
GRU and CNN are comparable when lengths are small, e.g., $<$10, then GRU gets increasing advantage over CNN when
meet longer sentences.
Error analysis shows that long sentences in SST mostly consist of clauses of inverse semantic such as ``this version
is not classic like its predecessor, but its pleasures are still plentiful''. This kind of
clause often include a local strong indicator for one sentiment polarity, like ``is not'' above,
but the successful classification relies on the comprehension of the whole clause.

Hence, \emph{which DNN type
performs better in text classification task depends on how often the comprehension of
global/long-range semantics is required}.

This can also explain the phenomenon in SemMatch -- GRU/LSTM
surpass CNN in TE while CNN dominates in AS, as textual
entailment relies on the comprehension of the whole
sentence \cite{bowman2015large}, question-answer in AS instead can be effectively
identified by key-phrase matching \cite{YinSXZ16}.

\paragraph{Sensitivity to hyperparameters}
We next check how stable the performance of CNN and GRU are
when hyperparameter values are varied.
\figref{senti} shows the performance of
CNN, GRU and LSTM for different learning rates, hidden sizes
and batch sizes.
All models are relativly smooth with respect to learning rate changes.
In contrast, variation in hidden size and batch
size cause large oscillations. Nevertheless, we still
can observe that CNN curve is mostly below the curves of GRU
and LSTM in \emph{SentiC} task, contrarily located at the
higher place in \emph{AS} task.

\section{Conclusions}
This work compared the three most widely used DNNs -- CNN,
GRU and LSTM -- in representative sample of NLP tasks. We found that
RNNs perform well and robust in a broad range of tasks except when
the task is essentially a  keyphrase recognition task as in
some sentiment detection and question-answer matching
settings. In addition, hidden size and batch size can make DNN
performance vary dramatically. This suggests that
optimization
of these two parameters is crucial to good performance of
both CNNs and RNNs.

\bibliography{acl2017}
\bibliographystyle{acl_natbib}
\end{document}